\pdfoutput=1
\documentclass[11pt]{article}

\usepackage[preprint]{acl}

\usepackage{times}
\usepackage{latexsym}

\usepackage[T1]{fontenc}
\usepackage[utf8]{inputenc}

\usepackage{microtype}

\usepackage{todonotes}

\usepackage{inconsolata}

\usepackage{algorithm}
\usepackage{algpseudocode}
\usepackage{graphicx}
\usepackage{amsmath}
\usepackage{amssymb}
\usepackage{booktabs}
\usepackage{multirow}
\usepackage{makecell}
\usepackage{tabularx}
\usepackage{times}
\usepackage{booktabs}
\usepackage{latexsym}
\usepackage{graphicx}
\usepackage{multirow}
\usepackage{multicol}
\usepackage{hyperref}
\usepackage{listings}
\usepackage{amsmath}
\usepackage[most]{tcolorbox}
\usepackage{float}
\usepackage{xspace}
\usepackage{graphicx}
\usepackage{subcaption}
\usepackage{musicography}

\title{Agentic-R1: Distilled Dual-Strategy Reasoning}

\author{
  Weihua Du\quad Pranjal Aggarwal\quad Sean Welleck\quad Yiming Yang \\
  Language Technologies Institute, Carnegie Mellon University \\
  \texttt{\{weihuad, pranjala, swelleck, yiming\}@cs.cmu.edu}
}%\\\And
\begin{document}
\maketitle
\begin{abstract}
%Current long chain-of-thought (long-CoT) models excel at mathematical reasoning but depend on slow and error-prone natural language traces. Tool-augmented agents address arithmetic through code execution but often struggle with complex logical reasoning. We introduce \textit{DualDistill}, a fine-tuning framework that distills complementary strategies from multiple teacher models into a unified student. Using this method, we train \textbf{Agentic-R1}, which learns to dynamically choose the optimal strategy per query, invoking tools for arithmetic tasks and relying on text-based reasoning for more abstract problems. Our approach improves performance on tasks requiring intensive computation and reduces inference latency on standard benchmarks, demonstrating the effectiveness of multi-strategy distillation for robust and efficient reasoning.

Current long chain-of-thought (long-CoT) models excel at mathematical reasoning but rely on slow and error-prone natural language traces. Tool-augmented agents address arithmetic via code execution, but often falter on complex logical tasks. We introduce a fine-tuning framework, \textbf{DualDistill}, that distills complementary reasoning strategies from multiple teachers into a unified student model. Using this approach, we train \textbf{Agentic-R1}, which dynamically selects the optimal strategy for each query, invoking tools for arithmetic and algorithmic problems, and using text-based reasoning for abstract ones. Our method improves accuracy across a range of tasks, including both computation-intensive and standard benchmarks, demonstrating the effectiveness of multi-strategy distillation in achieving robust and efficient reasoning. Our project is available at \href{https://github.com/StigLidu/DualDistill}{https://github.com/StigLidu/DualDistill}.
\end{abstract}
\section{Introduction}

A recently proposed reasoning paradigm for language models, long chain-of-thought (long-CoT) reasoning, has achieved state-of-the-art performance on challenging tasks such as mathematical problem solving~\cite{guo2025deepseek, jaech2024openai}. By allocating a large inference budget, these models generate reasoning trajectories with iterative self-verification and refinement. Despite this progress, open-source long-CoT models remain limited: Their reasoning traces rely solely on natural language, which is both computationally expensive and error-prone without explicit verification.

\begin{figure}[ht]
    \centering
    \includegraphics[width=0.482\textwidth]{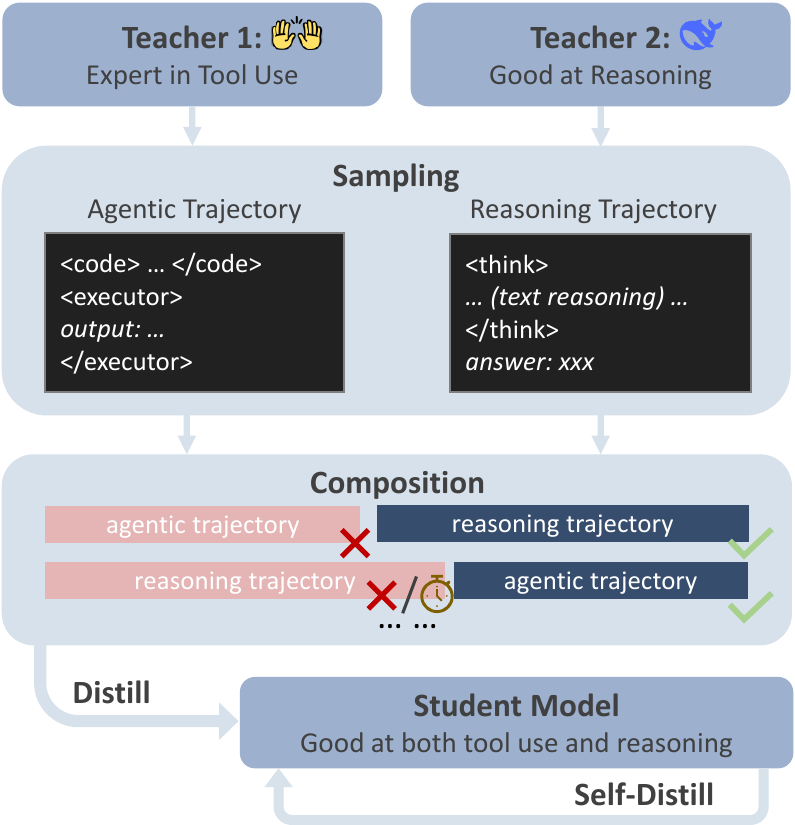}
    \vspace{-4mm}
    \caption{\textbf{Overview of \textit{DualDistill}.} We distill knowledge from two complementary teacher models. Trajectories from teachers are composed based on correctness, enabling the student model to learn when and how to select the appropriate strategy for each problem. Furthermore, the student internalizes these strategies through self-distillation.}
    \label{fig: intro}
    \vspace{-2mm}
\end{figure}

In contrast, tool-aided reasoning provides greater efficiency and reliability, particularly for large-scale numerical computations and tasks that require rigorous verification~\cite{gao2023pal}. Advanced agent frameworks, such as OpenHands~\cite{wang2024openhands}, place language models in a multi-turn environment with a code interpreter and other tools. The resulting agentic trajectories %(see Fig.\ref{fig.example_agentic_trajectory}) 
are effective for tool-intensive tasks, but often fall short on abstract or conceptually complex reasoning problems~\cite{duan2024gtbench}.

To leverage the strengths of both reasoning and tool-based strategies, we introduce \textbf{DualDistill}, a novel distillation framework (Fig.~\ref{fig: intro}) that combines trajectories from two complementary teachers: one reasoning-oriented, the other tool-augmented, in a unified student. The resulting model, \textbf{Agentic-R1}, learns to mix both strategies and dynamically selects the most appropriate one for each problem, executing code for arithmetic and algorithmic tasks and reasoning in natural language for abstract ones. Furthermore, the student can continue to improve via self-distillation, better calibrating its strategy selection to its actual capabilities. Our contributions are as follows.
\begin{itemize}
\item \textbf{DualDistill}, a distillation framework that enables a language model to learn from multiple teacher models with complementary capabilities through solution trajectory composition.
\item \textbf{Agentic-R1}, a distilled student model that achieves strong performance in mathematical tasks requiring both tool use and complex reasoning.
\end{itemize}
%\ifnum\ifarxiv=1
%\input{appendix/related}
%\fi
%\ifnum\ifarxiv=0
\section{Related Work}
\label{sec:related_work}
Although prior efforts have integrated external tools into language models~\cite{gao2023pal,schick2023toolformerlanguagemodelsteach,nakano2022webgptbrowserassistedquestionansweringhuman}, they are often specialized to non-math domains or are confined to shorter reasoning chains. Concurrently, the paradigm of long chain-of-thought (long-CoT) reasoning or inference-time compute has demonstrated significant improvements~\cite{guo2025deepseek,jaech2024openai}. However, these approaches can be difficult to control and may suffer from `overthinking', particularly when applied to tool-use scenarios~\cite{cuadron2025dangeroverthinkingexaminingreasoningaction}. Recent works have combined tool use with long-CoT reasoning~\cite{feng2025retoolreinforcementlearningstrategic,song2025r1searcherincentivizingsearchcapability}, but these are often applied to different domains or rely on reinforcement learning, which can be less stable than our proposed distillation method. To the best of our knowledge, \textbf{DualDistill} is the first framework to employ distillation with trajectory composition from two heterogeneous teacher models, one specializing in agentic tool-use and the other in pure textual reasoning, creating a unified student model capable of adaptively leveraging both strategies. See Appendix~\ref{app.related_work} for a more detailed discussion.
%\fi

%\section{DualDistill}

\section{Method}

%Recent advances in language models have resulted in specialized frameworks tuned to particular problem-solving domains, such as textual reasoning or software-based agentic systems. Although these domain-specific models individually demonstrate strong performance, combining their complementary strengths can further enhance general problem-solving capabilities. To this end, we propose \textit{DualDistill}, a distillation scheme with trajectory composition (Section~\ref{sec:data_composition}), enabling a unified student model to take advantage of multiple domains. Furthermore, we introduce a targeted problem-filtering procedure (Section~\ref{sec.problem_filter}) to select training problems that highlight clear distinctions between strategies.

As illustrated in Fig.~\ref{fig: intro}, \textbf{DualDistill} uses trajectory composition to distill the knowledge of the complementary teachers to the student model. The student model then applies self-distillation for a deeper understanding of the strategies.

\subsection{Trajectory Composition}
\label{sec:data_composition}

%Consider two distinct teacher policies: an \textit{agentic} teacher with policy $\pi_A$ and a \textit{reasoning} teacher with policy $\pi_R$. Given a training dataset $\mathcal{D}=\{(x_i, a_i)\}$, where $x_i$ denotes the $i$-th problem and $a_i$ is its ground-truth solution. The agentic approach more effectively solves some problems in $\mathcal{D}$, while others benefit from pure reasoning.

%To exploit the strengths of both teachers, we introduce a mixture policy $\pi_{\text{mix}}$. For each training instance $(x, a)$, we randomly determine the initial teacher by sampling a binary indicator $z \sim \text{Bernoulli}(0.5)$. The selected teacher then produces an initial solution $y_1$:

Let $\mathcal{D} = \{(x^{(i)}, a^{(i)})\}_{i=1}^N$ be a training set, where $x^{(i)}$ denotes the $i$-th problem and $a^{(i)}$ is its corresponding reference answer. Let $\pi_A$ and $\pi_R$ be two distinct teacher policies, where $\pi_A$ represents the \textit{agentic} teacher and $\pi_R$ the \textit{reasoning} teacher.
For each training instance $(x, a)$, we randomly select the initial teacher by sampling a binary indicator $z \sim \text{Bernoulli}(0.5)$ 
and then produce solutions $y_1$ and $y_2$ as follows:
\begin{align*}
y_1 &\sim z \pi_A(\cdot \mid x) + (1 - z) \pi_R(\cdot \mid x), \\
y_2 &\sim (1 - z) \pi_A(\cdot \mid x, y_1) + z \pi_R(\cdot \mid x, y_1).
\end{align*}
That is, after one teacher generates the initial solution $y_1$, the other teacher subsequently generates the second solution $y_2$, conditioned on both the original problem $x$ and the preceding solution $y_1$.

We evaluate the correctness of each solution using a rule-based grader, assigning binary correctness scores $g_1, g_2 \in \{0,1\}$ to $y_1$ and $y_2$, respectively. The distilled training trajectories are then composed based on these correctness scores.
\begin{itemize}
\item \textbf{$g_1 = 0, g_2 = 1$}: The first teacher produces an incorrect solution, and the second teacher successfully corrects it. The composed trajectory is structured as $y_1 \oplus t_{-+} \oplus y_2$. (Here $\oplus$ denotes concatenation and $t_{-+}$ is a transition segment, described later).
\item \textbf{$g_1 = 1, g_2 = 1$}: Both teachers provide correct solutions. We create a trajectory $y_1 \oplus t_{++} \oplus y_2$ to reflect complementary correct strategies.
\item \textbf{$g_1 = 1, g_2 = 0$}: Only the initial teacher provides a correct solution. In this scenario, the composed trajectory includes only $y_1$.
\item \textbf{$g_1 = 0, g_2 = 0$}: Both teachers do not solve the problem correctly. %Such instances are discarded and excluded from the training dataset.
In this case, we just discard the problem without composing any trajectory.
\end{itemize}
The transition segments $t_{-+}$ and $t_{++}$ are predefined sentences indicating strategy shifts (e.g., \textit{"Wait, using text reasoning is too tedious, let us try code reasoning."}). More examples and detailed descriptions can be found in Appendix~\ref{app.data_example}.

\subsection{Training Instance Selection}
\label{sec.problem_filter}
%To effectively train a student model capable of switching between agentic and pure reasoning strategies, 
We curate a training set with the instances for which one strategy has a clear advantage over the other in performance. Using an existing dataset such as GSM8K~\cite{cobbe2021training} would be insufficient in this sense
%each strategy shows a clear advantage in different problem types. A straightforward dataset like GSM8K~\cite{cobbe2021training} is insufficient for this purpose, 
as most of the problems are relatively simple and can be solved by either strategy without a significant performance difference. Instead, we construct two contrasting subsets of Math problems from DeepMath~\cite{he2025deepmath}: one can benefit more from tool-assisted reasoning, while the other can benefit more from pure text-based reasoning. After composition,
%one where tool-assisted reasoning provides substantial benefit, and another where pure reasoning outperforms tool use. This enables the student model to learn more precise decision boundaries between strategies. 
we apply additional filtering to balance the training dataset, resulting in $2.6$k distilled trajectories. Detailed statistics can be found in Appendix~\ref{app.dataset_scale}.
Further filtering details are provided in the Appendix~\ref{appendix.problem_filter}.

%\subsection{Implementation Details}
\subsection{Teacher and Student Models}
%\paragraph{Teacher Model Policy.} We select two representative teacher models corresponding to agentic and reasoning strategies. 
As the teacher of agentic reasoning, we utilize OpenHands~\cite{wang2024openhands}, a tool-assisted agent built upon \textit{Claude-3.5-Sonnet}~\cite{anthropic2024claude35sonnet} %framework 
to employ human-designed problem-solving pipelines.  %For the reasoning strategy, 
As the teacher of text-based reasoning, we adopt \textit{DeepSeek-R1}~\cite{guo2025deepseek}.
The details %The details of the trajectory composition of the two teacher models
can be found in Appendix~\ref{app.trajectory_composition_implementation}.

%\paragraph{Student Model.} W
As for the student model, we adopt \textit{DeepSeek-R1-Distill-7B}, which %as our student model. %This model
has been fine-tuned on pure text-based reasoning trajectories and also exposed to code-related data during pretraining. %Our goal is to examine whether it can effectively integrate multiple problem-solving strategies through composition distillation.
We deliberately choose a model already familiar with both modalities to minimize the amount of training data required for the strategic composition.
We want to examine whether it can effectively learn %integrate 
multiple problem-solving strategies.% through composition distillation. 
%Training a model to acquire an entirely new strategy from scratch typically requires significantly more data, which we leave as an avenue for future work (see Section~\ref{sec.limitaions} for further discussion).

\algnewcommand\algorithmicswitch{\textbf{switch}}
\algnewcommand\algorithmicendswitch{\textbf{end switch}}

\algdef{SE}[SW]{Switch}{EndSwitch}[1]%
  {\algorithmicswitch\ #1}%
  {\algorithmicendswitch}

\begin{algorithm}[ht]
\caption{\textsc{DualDistill}}
\label{algo:dual_distill}
\begin{algorithmic}[1]
\State \textbf{Input:} Teacher policies $\pi_A,\pi_R$; student $S_0$; training dataset $\mathcal{D} = \{(x_i, a_i)\}_{i=1}^N$; thresholds $\beta_1,\beta_2$; sample count $K$; binary grader $G(\cdot,\cdot)$
\State \textbf{Output:} Trained student $S_2$
\Statex \textcolor{gray}{\sc Teacher Distillation}
\State Initialize teacher-distillation buffer $\mathcal T_1 \gets \varnothing$
\For{each $(x,a)\in\mathcal{D}$}
      \State Draw $z\sim\mathrm{Bernoulli}(0.5)$
      \State $y_1 \sim z\,\pi_A(\cdot\!\mid\!x) + (1{-}z)\,\pi_R(\cdot\!\mid\!x)$
      \State $y_2 \sim (1{-}z)\,\pi_A(\cdot\!\mid\!x,y_1) + z\,\pi_R(\cdot\!\mid\!x,y_1)$
      \State $g_1\!\gets\!G(y_1,a),\;g_2\!\gets\!G(y_2,a)$
    \Switch{$(g_1,g_2)$}
      \State \textbf{case $(0,1)$}: Add $y_1 \oplus t^{-+} \oplus y_2$ to $\mathcal{T}_1$
      \State \textbf{case $(1,1)$}: Add $y_1 \oplus t^{++} \oplus y_2$ to $\mathcal{T}_1$
      \State \textbf{case $(1,0)$}: Add $y_1$ to $\mathcal{T}_1$
    \EndSwitch
\EndFor
\State Balance $\mathcal{T}_1$
\State Fine-tune $S_0$ on $\mathcal T_1 \rightarrow S_1$
\Statex \textcolor{gray}{\sc Self-Distillation}
\State Initialize self-distillation buffer $\mathcal T_2 \gets \varnothing$
\For{each $(x,a)\in\mathcal{D}$}
  \State Sample $\{t_j\}_{j=1}^K \sim \pi_{S_1}(\cdot\!\mid\!x)$
  \State $g_j \gets G(t_j,a)$
  \State $\bar g \gets \frac1K\sum_{j=1}^{K} g_j$
  \If{$\bar g > \beta_1$}
    \State Add a correct $t_j$ + verification to $\mathcal T_2$
  \EndIf
  \If{$\bar g < \beta_2$}
    \State Add an incorrect $t_j$ + correction to $\mathcal T_2$
  \EndIf
\EndFor
\State Fine-tune $S_1$ on $\mathcal T_2 \rightarrow S_2$
\State \Return $S_2$
\end{algorithmic}
\end{algorithm}
\subsection{Self-Distillation}
Although the student model learns problem-solving strategies from multiple teachers, it can still underperform compared to them due to limitations such as the smaller model sizes, leading to an ability mismatch.
%as shown in Table~\ref{tab:main_result}, after distillation from the teacher models, our trained model \textit{Agentic-R1-7B} performs slightly worse on certain datasets where it initially demonstrated strong performance (\textit{e.g.,} MATH500 and AMC). 
For instance, we find that the student sometimes uses tools for problems that could be solved more reliably through simple reasoning. While this approach is valid, it can introduce errors because the student's tool-use ability is less mature than that of the teachers, occasionally leading to incorrect code and wrong answers.

To address it, we introduce \textit{self-distillation} to help the student further refine its strategy selection based on its capabilities and the given problem. Our self-distillation process involves the student model generating candidate solutions, with teacher models providing verification or corrections as auxiliary supervision. The process reinforces effective strategies and corrects suboptimal ones. Specifically, given a training set $\mathcal{D} = \{(x^{(i)}, a^{(i)})\}_{i=1}^N$ and the student policy fine-tuned on distillation data from teachers $\pi_{S_1}$, we sample $K$ trajectories $t^{(i,1)}, \dots, t^{(i,K)}$ for each problem $x^{(i)}$:
$$
t^{(i,j)} \sim \pi_{S_1}(\cdot \mid x^{(i)}), \; x^{(i)} \in \mathcal{D}, \; j \in \{1, \dots, K\}.
$$
We then apply a binary grader $G$ to evaluate trajectory accuracy. Let $g^{(i,j)}$ be the score of the $j$-th trajectory and $\overline{g}^{(i)}$ be the average score for $x^{(i)}$, i.e.,
$$
g^{(i,j)} = G(t^{(i,j)}, a^{(i)}), \; \overline{g}^{(i)} = \frac{1}{K} \sum_{j=1}^{K} g^{(i,j)}.
$$
If $\overline{g}^{(i)} \ne 1$, the student cannot fully solve problem $x^{(i)}$, and we collect informative trajectories from its output to form a self-distillation buffer for further training. Specifically:
\begin{itemize}
\item If $\overline{g}^{(i)} > \beta_1$, we add a correct trajectory generated by the student, followed by a verification from a teacher model, to the replay buffer;
\item If $\overline{g}^{(i)} < \beta_2$, we add an incorrect student trajectory, along with a corrected solution provided by a teacher, to the buffer.
\end{itemize}
Here, $\beta_1$ and $\beta_2$ are hyperparameters that control the difficulty of the problems selected for the replay buffer. We set $\beta_1 = 0$ and $\beta_2 = 0.9$, a relatively low threshold that encourages diversity in the collected examples. In addition, we use $K = 16$ trajectory samples per problem. Verification (or correction) consists of a correct trajectory from the teacher model with some transition words; see Appendix \ref{app.data_example} for details. Because we observed a gap in the coding ability between the student and the teacher, we provide only text-based reasoning solutions as the teacher’s answers at this stage.

The pseudocode for our complete algorithm, \textbf{DualDistill}, is listed in Algorithm~\ref{algo:dual_distill}.
\section{Experiments}
\begin{table*}[ht]
    \centering
    \small
\begin{tabular}{l|c|ccccc|c}
\toprule
\multicolumn{1}{c|}{Model} & Budget & DeepMath-L  & Combinatorics300 & MATH500 & AIME & AMC & avg.\\
\midrule
Qwen2.5-7B-Instruct (w/o tool) 
    & \makecell{S\\L}
    & \makecell{17.2\\17.5}
    & \makecell{21.8\\21.8}
    & \makecell{75.1\\75.2}
    & \makecell{8.0\\8.0}
    & \makecell{42.9\\42.9} 
    & \makecell{33.0\\33.1}
    \\
\midrule
Qwen2.5-7B-Instruct (w/ tool)    
    & \makecell{S\\L}
    & \makecell{34.7\\34.7}
    & \makecell{28.9\\28.9}
    & \makecell{70.2\\70.2}
    & \makecell{14.7\\14.7}
    & \makecell{51.1\\51.1}
    & \makecell{39.9\\39.9}
    \\
\midrule
DeepSeek-R1-Distill-7B 
    & \makecell{S\\L}
    & \makecell{34.7\\56.3}
    & \makecell{34.7\\44.5}
    & \makecell{\textbf{83.1}\\\underline{89.2}}
    & \makecell{23.3\\\textbf{40.7}}
    & \makecell{61.2\\\underline{84.8}}
    & \makecell{47.4\\\underline{63.1}}
    \\
\midrule
Agentic-R1-7B (ours) 
    & \makecell{S\\L}
    & \makecell{\underline{37.0}\\\underline{59.3}}
    & \makecell{\underline{36.9}\\\underline{49.4}}
    & \makecell{80.0\\82.4}
    & \makecell{\textbf{28.0}\\\textbf{40.7}}
    & \makecell{\underline{64.3}\\82.2}
    & \makecell{\underline{49.3}\\62.8} 
    \\
\midrule
Agentic-R1-7B-SD (ours)
    & \makecell{S \\L}
    & \makecell{\textbf{40.0}\\\textbf{65.3}}
    & \makecell{\textbf{38.2}\\\textbf{52.0}}
    & \makecell{\underline{82.5}\\\textbf{93.3}}
    & \makecell{\underline{27.3}\\\textbf{40.7}}
    & \makecell{\textbf{66.3}\\\textbf{85.8}}
    & \makecell{\textbf{50.9}\\\textbf{67.4}}
    \\
\bottomrule
\end{tabular}
\caption{\textbf{Main Results.} We evaluate on five benchmarks under two budgets: \textbf{S} (4096) and \textbf{L} (32768). The results are averaged over 5 seeds with $T=0.6$. The best results are highlighted in bold, and the second-best results are underlined. \textit{Agentic-R1} demonstrates significant gains on DeepMath-L and Combinatorics300, where both complex reasoning and tool use are crucial, while maintaining comparable performance on common math tasks. Furthermore, through self-distillation, \textit{Agentic-R1-SD} can enhance performance and outperform baselines on nearly all tasks.}
\label{tab:main_result}
\end{table*}
\subsection{Benchmarks}

We evaluated our method on several benchmarks that test different aspects of mathematical reasoning, including tasks where tool-aided calculation is hypothesized to provide a significant advantage.

\paragraph{DeepMath-L.}
DeepMath~\cite{he2025deepmath} is a comprehensive dataset of mathematical and STEM problems compiled from various benchmarks. We curate a subset of 87 problems with large answers (absolute value greater than $10^5$). These problems are excluded from our fine-tuning data, although they may appear in some pretraining corpora. We refer to this evaluation set as \textit{DeepMath-L}, with the assumption that code-aided computation is more effective in solving such problems.

\paragraph{Combinatorics300.}
This benchmark consists of 300 combinatorics problems aggregated from diverse math test sets. Each problem yields an answer larger than $10^4$, reflecting the factorial growth in combinatorial counts. We hypothesize that tool-aided reasoning is important for handling the enumeration and sampling required in such tasks.

\paragraph{Standard Mathematical Benchmarks.}
To assess the generalizability of our approach, we further evaluate on widely used mathematical reasoning tasks, including MATH500~\cite{lightman2023let}, AMC~\cite{amc2024}, and AIME (2025, Parts I and II)~\cite{aime2025}.

\subsection{Baselines}

We compare against the following strong baselines:
\begin{itemize}
\item \textbf{DeepSeek-R1-Distill.} A distilled version of DeepSeek-R1 fine-tuned on long chain-of-thought trajectories, representing a strong baseline for pure language-based reasoning.
\item \textbf{Qwen-2.5-Instruct (w/ tool, w/o tool)}~\cite{yang2024qwen2}. A general-purpose short-CoT model with optional tool-use capabilities. The tool-augmented variant serves as a competitive baseline for tool-aided strategies.
\end{itemize}
The training configuration details are provided in Appendix~\ref{app.train_configuration}.

\subsection{Evaluation Metrics}
To assess both reasoning quality and computational efficiency, we adopt the \textit{Accuracy at Budget} metric. Let $t = (t_0, t_1, \dots, t_L)$ be the trajectory generated by the model, where each $t_\ell$ denotes the $\ell$-th output token, and let $a$ be the reference answer. The accuracy under a computational budget $b$ is defined as:
$$
\operatorname{Acc}(b) = \textsc{G}\big(t_{0:\min(b, L)},\, a\big),
$$
where \textsc{G} is a binary grader that evaluates whether the model's partial output matches the ground truth. We report results under two budgets: \textit{Standard (S, 4096)}, a moderate token budget for language model reasoning, and \textit{Large (L, 32768)}, which approximates an unbounded budget and allows the model to reason adequately. Inference and grader details can be found in Appendix~\ref{app.inference_details}.

\subsection{Results}
As shown in Table~\ref{tab:main_result}, our student model, \textit{Agentic-R1}, demonstrates substantial performance improvements in DeepMath-L and Combinatorics300, two challenging datasets that benefit from both agentic and reasoning strategies. Specifically, our model outperforms two similarly sized models, each specializing exclusively in tool-assisted (\textit{Qwen2.5-7B-Instruct}) or pure reasoning (\textit{DeepSeek-R1-Distill-7B}) strategies. \textit{Agentic-R1} surpasses tool-based models by intelligently adopting reasoning strategies when appropriate, while maintaining greater efficiency compared to pure reasoning models on standard mathematical tasks. However, we note a slight performance decrease in relatively simpler benchmarks (MATH500) compared to the pure text-reasoning model, and a detailed discussion is provided in the limitations section.

\paragraph{Qualitative Examples.}
We provide illustrative trajectories demonstrating \textit{Agentic-R1}’s adaptive strategy-switching capability: (1) initially using the tool-assisted strategy and then switching to textual reasoning to correct an incorrect initial solution (Fig.~\ref{fig.agentic-r1_behavior_1}); and (2) starting with textual reasoning and then switching to the tool-assisted strategy to bypass tedious manual calculations (Fig.~\ref{fig.agentic-r1_behavior_2}).

\paragraph{Agentic-R1 Knows When to Use Tools.}
An intriguing observation is that \textit{Agentic-R1} learns when to appropriately invoke code tools purely through supervised fine-tuning. For instance, Combinatorics300 contains problems involving large numerical computations, which makes the tools particularly beneficial. Consequently, \textit{Agentic-R1} activates at least one code execution tool in $79.2\%$ of the Combinatorics300 problems, while the usage of the tool drops to only $52.0\%$ in the relatively simpler AMC dataset.

\paragraph{Agentic-R1 Learns from Imperfect Teachers.}
Although OpenHands, based on \textit{Claude-3.5-Sonnet}, is not a strong standalone reasoning agent and sometimes performs worse than the student’s initial model (\textit{R1-Distill-7B}), the student model still effectively acquires valuable agentic strategies through distillation. For example, the agentic strategy teacher achieves only $48.4\%$ accuracy on Combinatorics300, yet after training, the student's performance improves significantly from $44.7\%$ to $50.9\%$, surpassing the teacher. 
This shows that demonstrations from an imperfect agentic teacher can still yield meaningful gains in the student.
% Notably, the student initially possessed strong capabilities in pure textual reasoning, underscoring the effectiveness of learning even from imperfect demonstrations.

%\paragraph{Will Code Executor Introduce Extra Cost?}
%\weihua{Relatively cheap O(1)S, compared with inference length O(L2)}
%maybe in appendix?

\subsection{Ablation Study}
\begin{table}[h]
    \centering
    \small
    \begin{tabular}{l|ccc}
    \toprule
    Dataset & DeepMath-L & AIME & AMC \\
    \midrule
    w/o composition & 40.0\% &  34.0\% & 50.8\% \\
    w/ composition & \textbf{59.3\%} & \textbf{40.7\%} & \textbf{82.2\%} \\
    \bottomrule
    \end{tabular}
    \caption{\textbf{Trajectory Composition}. We compare performance between composition and non-composition distillation in the large budget setting; composition is always better.}
    \label{tab:ablation}
\end{table}
\paragraph{Trajectory Composition.} To verify the effectiveness of our data composition strategy, we compare it with a training strategy that does not use composition, meaning that each student trajectory is either fully generated by the agentic teacher or fully generated by the reasoning teacher.
% only receives the correct parts of the trajectories. 
As shown in Table~\ref{tab:ablation}, our composition strategy consistently surpasses its non-composition counterpart.
\section{Conclusion}
\label{sec.conclusion}
We propose \textbf{DualDistill}, an efficient distillation framework based on trajectory composition, allowing a student model to learn from multiple teacher models specialized in different domains of problem solving. Using the appropriate strategy for each problem, our trained model, \textit{Agentic-R1}, achieves superior performance in benchmarks that require both reasoning and tool-assisted capabilities. This approach demonstrates the potential for unifying diverse problem-solving strategies within a single model, opening new directions for building versatile and adaptive language agents.

\section*{Limitations}
\label{sec.limitaions}
While our approach demonstrates strong overall performance, several limitations remain for future work. First, the transition words connecting different strategies within composed trajectories are currently designed manually. As a result, the output produced by the trained student model can occasionally lack naturalness and fluidity, especially when switching between strategies. Moreover, the student model after self-distillation may exhibit repetitive behavior. Developing methods for more coherent and automatic transitions between strategies could further enhance the readability and overall quality of the content generated by the student model.

%First, we observe a slight performance decrease of \textit{Agentic-R1} on relatively simple reasoning tasks such as MATH500 and AMC, compared to the baseline of pure text reasoning (\textit{DeepSeek-R1-Distill}). 
%We hypothesize two main reasons. These tasks are well-suited for pure text-based reasoning, with the baseline achieving over $84\%$ accuracy. In such cases, tool-aided reasoning offers limited additional benefit. For example, when using tool assistance, \textit{Qwen-2.5-7B-Instruct} achieves only around $68.4\%$ accuracy on \textit{MATH500}—a performance drop compared to the tool-free baseline, indicating that code execution may be less useful for relatively simple problems. (2) Our current training scheme relies solely on strategy distillation. Combining preference-based learning methods, such as expert iteration~\cite{polu2022formal} or DPO~\cite{rafailov2023direct} afterwards, may help the student model better select the appropriate reasoning strategy.

Second, our training dataset contains approximately $2.6$k composed trajectories. While this appears sufficient to teach a model that has been pre-trained on both text reasoning and code generation (e.g., \textit{DeepSeek-R1-Distill-7B}) to choose between strategies, it is likely insufficient for training a model to learn a new reasoning strategy from scratch. For example, \textit{DeepSeek-R1-Distill} was fine-tuned on over $800$k distilled examples to acquire long CoT reasoning capabilities. Expanding the dataset and covering a wider range of strategies will be an important direction for future research.

\section*{Acknowledgments}
This work was supported in part by the National Science Foundation under Grant Nos. DMS-2434614 and DMS-2502281.

\bibliography{custom}

\appendix
\definecolor{lightgray}{gray}{0.95}
\lstdefinestyle{prompt}{
    basicstyle=\ttfamily\fontsize{7pt}{8pt}\selectfont,
    frame=none,
    breaklines=true,
    backgroundcolor=\color{lightgray},
    breakatwhitespace=true,
    breakindent=0pt,
    escapeinside={(*@}{@*)},
    numbers=none,
    numbersep=5pt,
    xleftmargin=5pt,
}
\tcbset{
  aibox/.style={
    top=10pt,
    colback=white,
    colframe=black,
    colbacktitle=black,
    enhanced,
    center,
    % breakable,
    attach boxed title to top left={yshift=-0.1in,xshift=0.15in},
    boxed title style={boxrule=0pt,colframe=white,},
  }
}
\newtcolorbox{AIbox}[2][]{aibox, title=#2,#1}

\section{Appendix}
\begin{figure*}[!ht] 
\vspace{-5mm}
\begin{AIbox}{Inference Prompt}
\small
{\color{blue}\bf System:}
{
A conversation between User and Assistant. The user asks a question, and the Assistant solves it.

The assistant first thinks about the reasoning process in the mind and then provides the user with the answer. 

The reasoning process and answer are enclosed within <think> </think> and <answer> </answer> tags, respectively, i.e., <think> reasoning process here </think> <answer> answer here </answer>.

The final answer should be enclosed within boxed tags, i.e., $\boxed{\text{answer here}}$.

{\color{brown}Meanwhile, you can use Python code to help you reason. The code should be enclosed within <code> </code> tags. For example, <code> code here </code>.

An executor will run the code and provide feedback immediately after the code. The executor feedback should be enclosed within <executor> </executor> tags.

You can use the executor feedback to improve your reasoning.}
}

\end{AIbox} 
\caption{\textbf{Inference Prompt.} The system prompt used to guide the model during inference. Instructions highlighted in {\color{brown} brown} indicate guidance specific to tool usage.}
\label{fig.prompt_template}
\vspace{-5mm}
\end{figure*}
\label{sec:appendix}

\subsection{Code and Dataset}
Our code is available at \href{https://github.com/StigLidu/DualDistill}{\url{https://github.com/StigLidu/DualDistill}}; Training data is available at \href{https://huggingface.co/datasets/VanishD/DualDistill}{\url{https://huggingface.co/datasets/VanishD/DualDistill}}.
\subsection{Training Configuration}
\label{app.train_configuration}
\paragraph{Loss Masking.}
To prevent the student model from learning incorrect patterns from unsuccessful attempts, we exclude specific segments of trajectories from the loss calculation. Specifically, trajectory segments occurring before a transition from incorrect to correct reasoning (i.e., $t_{-+}$) are omitted. Additionally, the executor's feedback and the code blocks resulting in execution errors are also excluded from influencing the loss computation.

\paragraph{Hyperparameters.}For fine-tuning the student model \textbf{Agentic-R1} on teacher distilled trajectories, we use 4$\times$ A6000 GPUs for a total of 12.7 hours. The model is trained for 4 epochs using the AdamW optimizer~\cite{loshchilov2017decoupled} with a learning rate of $1 \times 10^{-5}$. We set the maximum context length to 16,384 tokens for teacher distillation and 8,192 for self-distillation, and discard all training examples that exceed this limit.
\subsection{Dataset Details}
\subsubsection{Problem Filtering Heuristics}
\label{appendix.problem_filter}
To curate a training dataset that can guide a student model in learning when to apply agentic versus pure text-based reasoning, we construct two subsets of mathematical problems.

\paragraph{Agentic-Favored Subset.}
We identify problems where tool use is highly beneficial using two heuristics:
\begin{itemize}
    \item \textbf{Numerical Scale:} Problems whose final integer answers exceed an absolute value of $1,000$ often require nontrivial arithmetic operations or algorithms that are more suitable for tool-assisted computation.
    \item \textbf{Difficulty Under Constraints:} We use a baseline text reasoning-only model, \textit{DeepSeek-R1-Distill-7B}, with a limited context length (4096 tokens). Problems unsolvable under the constraint with one trial are deemed more difficult and suitable for agentic strategies.
\end{itemize}

\paragraph{Pure Reasoning-Favored Subset.}  
To balance the dataset, we include problems in which agent execution is error-prone. These are selected by identifying the cases where the tool-assisted strategy fails and produces incorrect output.

We apply this selection process to DeepMath-103K~\cite{he2025deepmath} and balance the two subsets to ensure that the model sees roughly equal representation from both strategies during training.
\subsubsection{Dataset Scale}
\label{app.dataset_scale}
\begin{table}[h]
    \centering
    \small
    \begin{tabular}{l|ccc}
    \toprule
    case & $g_1,g_2=1,1$ & $g_1,g_2=1,0$ & $g_1,g_2=0,1$ \\
    \midrule
    number & 685 & 600 & 1393 \\
    \bottomrule
    \end{tabular}
    \caption{\textbf{Dataset Scale}. We report the number of training examples in each correctness category. Recall that $g_1$ and $g_2$ represent the correctness of the first and second teachers, respectively.}
    \label{tab:dataset_scale}
\end{table}
After running the two teachers on the filtered subset and composing the trajectories, the final distilled dataset contains $2,678$ examples. The detailed count for each correctness category is listed in Tab.~\ref{tab:dataset_scale}.
\subsection{Composition Trajectory}
\subsubsection{Transition Segment}
\label{app.data_example}
When the teacher changes, a hand-designed transition segment is added to signify and point out the meaning of the transition. There are three typical transition segments $t$, which are shown in Table~\ref{tab.transition_segment}. The transition segments used in self-distillation are the same as those used in teacher distillation.
\newcolumntype{V}{>{\raggedright\ttfamily\arraybackslash}X}
\begin{table}[h!]
    \centering
    \small
    \begin{tabularx}{\linewidth}{c|V}
    \toprule
    Meaning & Content \\
    \midrule
    tool ($\times$) $\rightarrow$ text ($\checkmark$) & Wait, the code is not correct, let's try text reasoning.\\
    \midrule
    text ($\times$) $\rightarrow$ tool ($\checkmark$) & Wait, use text reasoning is too tedious, let's try code reasoning.\\
    \midrule
    A ($\checkmark$) $\rightarrow$ B ($\checkmark$) & Wait, we can also use \{B\}-reasoning as an alternative way to verify the solution.\\
    \bottomrule
    \end{tabularx}
    \caption{\textbf{Transition Segment.} The transition segments are used to connect trajectories from different teachers. `Tool' and `text' in the table represent agentic and pure text reasoning strategies, respectively. $\checkmark$ and $\times$ mean whether the trajectory is correct or not.}
    \label{tab.transition_segment}
\end{table}
\subsubsection{Trajectory Composition Implementation}
\label{app.trajectory_composition_implementation}
To transform multi-turn agentic trajectories from OpenHands logs into a suitable training format, we extract content from the log fields labeled `\textit{thought}', `\textit{code}', and `\textit{final thought}' along with their associated executor feedback, if any. Each extracted content is then enclosed within distinct resource tags—\textit{<think></think>}, \textit{<code></code>}, \textit{<answer></answer>} or \textit{<executor></executor>}—and concatenated sequentially. For reasoning trajectories from \textit{DeepSeek-R1}, we specifically apply the \textit{<answer></answer>} tag to the content outside the long CoT part (i.e., beyond the \textit{<think></think>} segment).

We aim for the student model to select the most efficient strategy inherently. Thus, we enforce a token budget on the first teacher's inference: If $y_1$ does not complete within a randomly determined inference budget $L_0$, the inference is stopped and labeled unsuccessful. In contrast, we do not impose any token budget constraint on the trajectory of the second teacher $y_2$.

During preliminary experiments, we observed substantial differences in the distributional characteristics between the OpenHands trajectories ($\pi_A$) and \textit{DeepSeek-R1} trajectories ($\pi_R$). To avoid performance degradation of $y_2$ due to potential contamination from combined inputs, we assume conditional independence and explicitly define the teacher model inference policy as $\pi(\cdot \mid x, y_1)=\pi(\cdot \mid x)$.
\subsection{Qualitative Example}
\label{app.qualitative_example}
We observed that \textbf{Agentic-R1} shows several promising behaviors: (1) The model initially adopts tool-aided reasoning, but yields incorrect outputs after several attempts, and then the model automatically switches to text reasoning and finally derives the correct answer (Fig.~\ref{fig.agentic-r1_behavior_1}); (2) The model initially tries to apply text reasoning for a combinatorial problem, and then changes to tool-aided reasoning to reduce computational complexity (Fig. ~\ref{fig.agentic-r1_behavior_2}).
\subsection{Inference Details}
\label{app.inference_details}
For all evaluation experiments, we use the VLLM framework~\cite{kwon2023efficient} to enable fast inference via prefix caching, which significantly accelerates multi-turn tool calls. In the tool-augmented setting, the language model is allowed to invoke a Python executor up to $10$ times per problem, with each execution capped at $3$ seconds. During inference, whenever the model outputs the special token \textit{</code>}, the generation process is temporarily paused, the preceding code block is executed, and the resulting feedback is appended to the ongoing generation enclosed with \textit{<executor> </executor>} before resuming inference. Although tool execution introduces up to $30$ seconds of additional runtime per query in the worst case, this cost is relatively small compared to the time-intensive pure text reasoning process, which can take several minutes to reach a conclusion using \textit{DeepSeek-R1-Distill-7B} on 2$\times$A6000 GPUs. Additionally, the prompt template is listed in Fig.~\ref{fig.prompt_template}.

The grader evaluates output trajectories in two steps:
\begin{itemize}
\item \textbf{Exact match}: The grader extracts the final non-think block: a code result (`<executor>…</executor>') or a direct answer (`<answer>…</answer>'), and compares it with the gold answer.
\item \textbf{Fuzzy match}: If no exact match is found, the full output is passed to \textit{MathVerify}~\cite{math_verify_2025}, an open-source verifier that checks mathematical equivalence. This helps capture correct answers that may appear earlier in the trace, ensuring a fairer comparison for long-CoT baselines (e.g., \textit{DeepSeek-R1-Distill}) when facing length truncation.
\end{itemize}
\subsection{Full Results}
\label{app.full_results}
\begin{figure}[ht]
    \centering
    \includegraphics[width=0.48\textwidth]{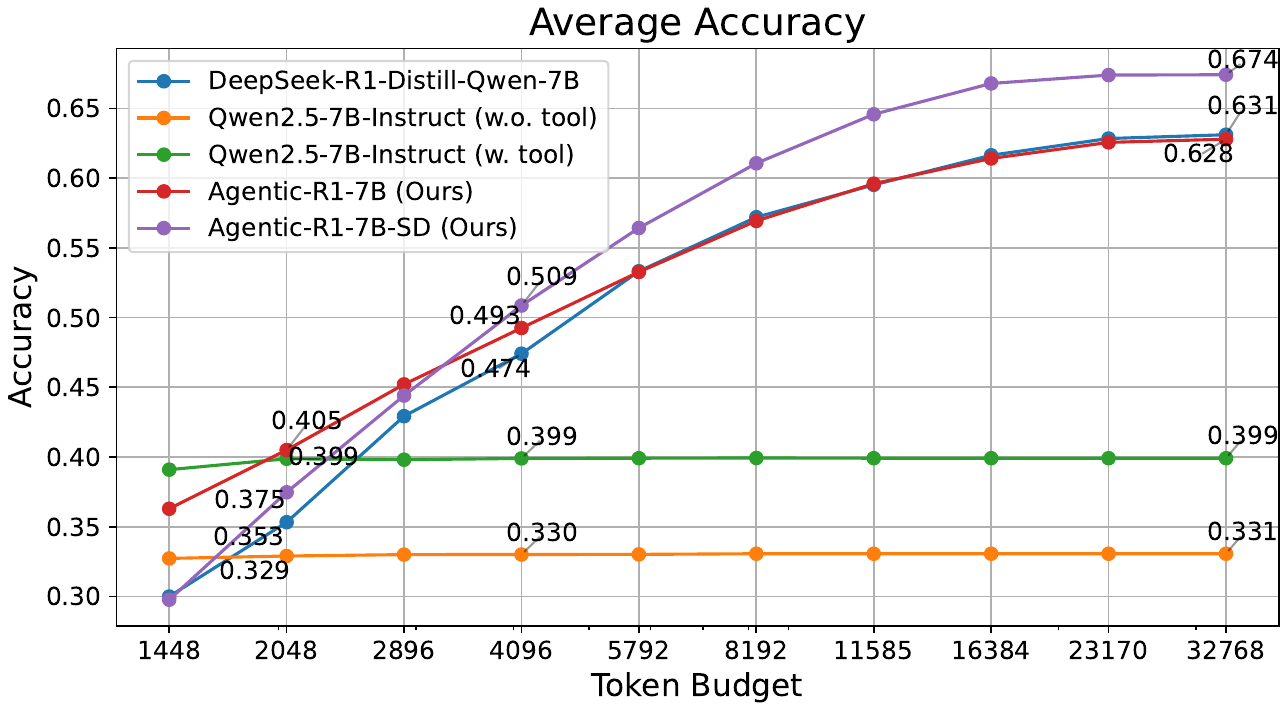}
    \vspace{-5mm}
    \caption{The average accuracy across benchmarks under various token budgets.}
    \label{fig.micro_avg_result}
\end{figure}

\begin{figure}[h!]
\centering
\begin{subfigure}{\linewidth}
    \centering
    \includegraphics[width=\linewidth]{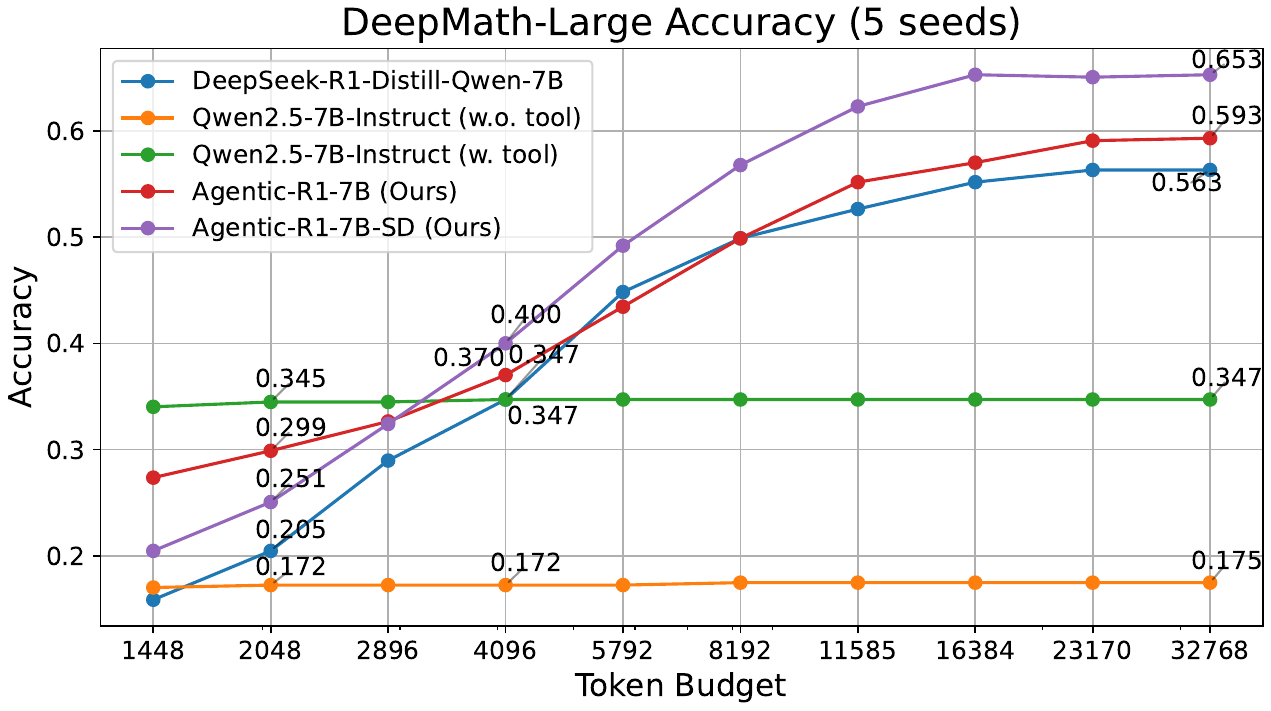}
\end{subfigure}

\vspace{2mm} % space between figures

\begin{subfigure}{\linewidth}
    \centering
    \includegraphics[width=\linewidth]{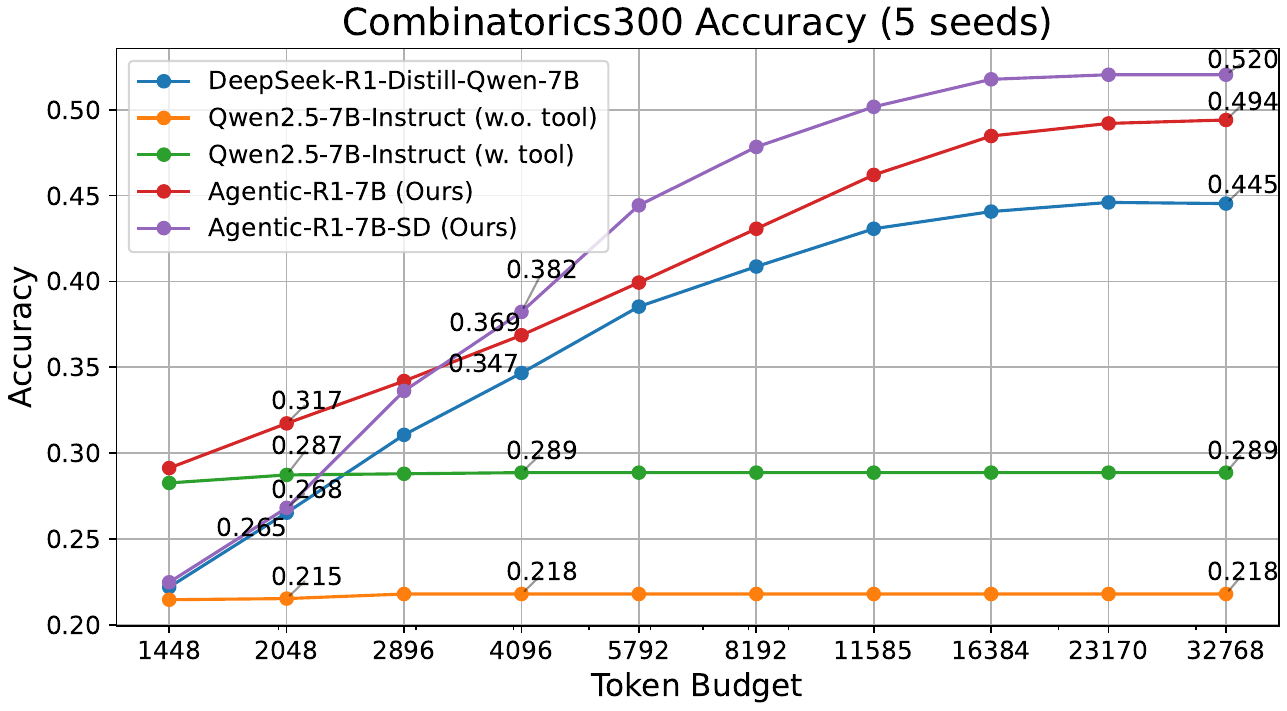}
\end{subfigure}

\vspace{2mm}

\begin{subfigure}{\linewidth}
    \centering
    \includegraphics[width=\linewidth]{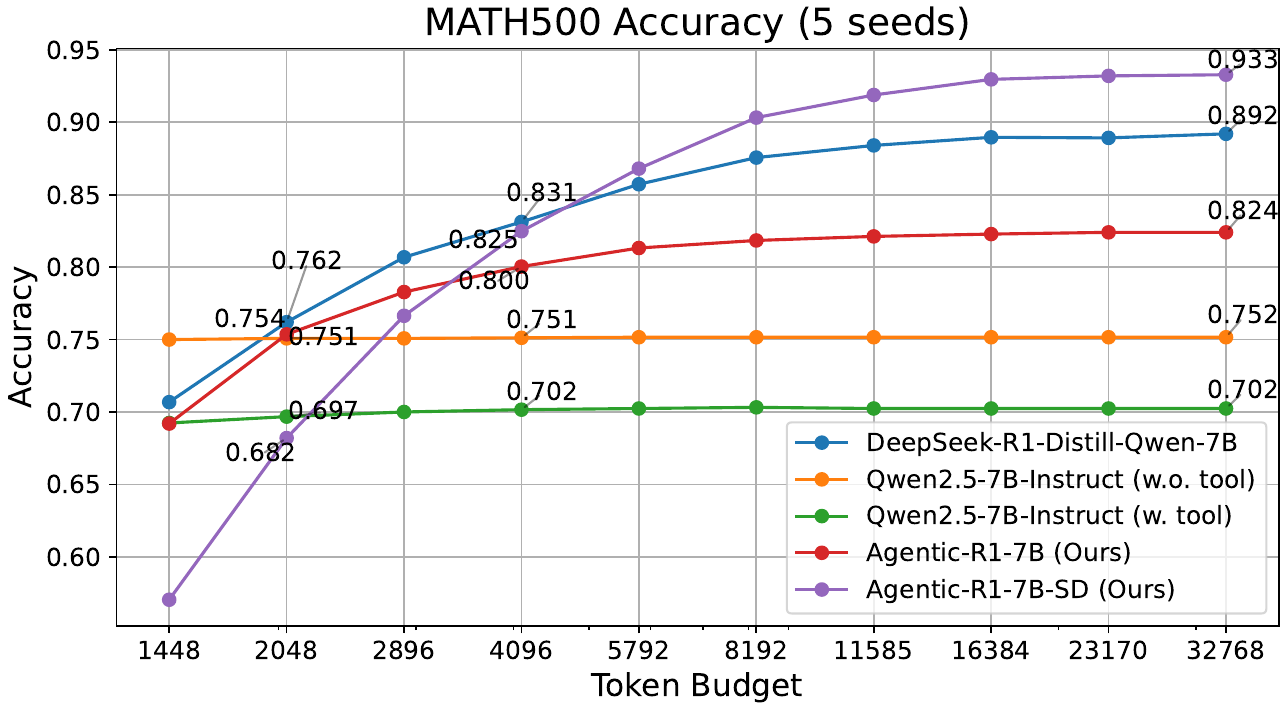}
\end{subfigure}

\vspace{2mm}

\begin{subfigure}{\linewidth}
    \centering
    \includegraphics[width=\linewidth]{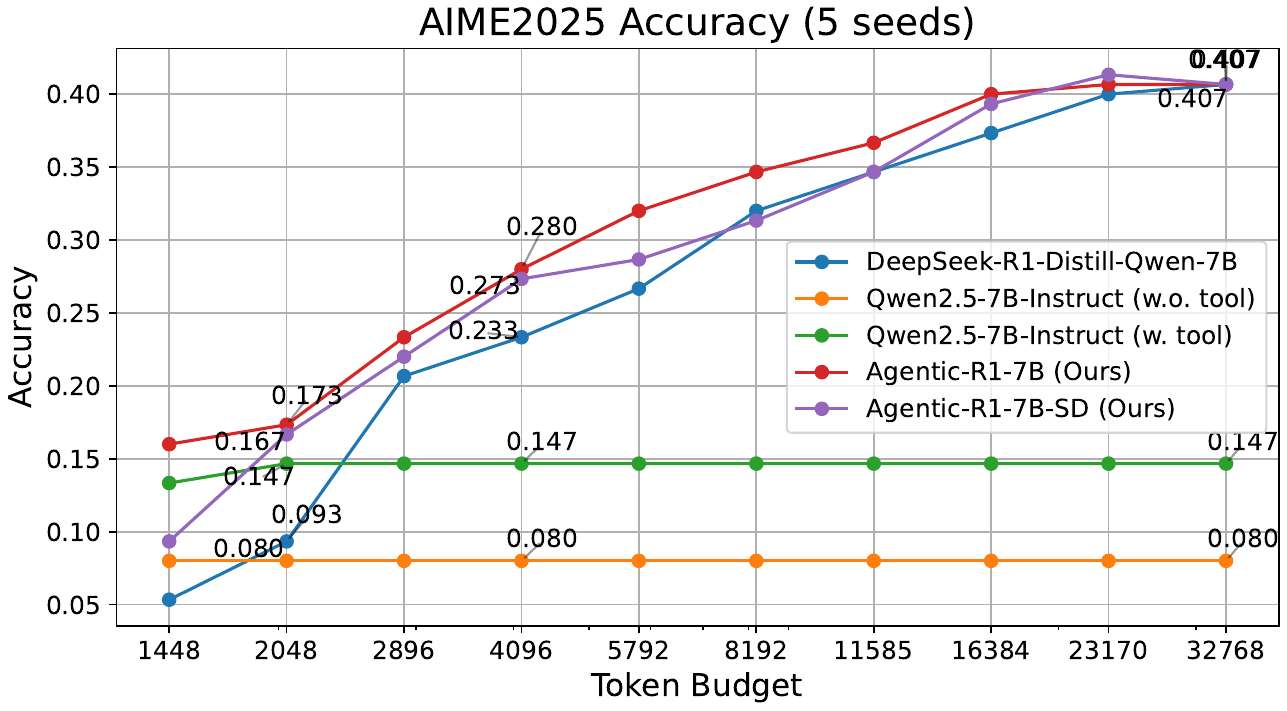}
\end{subfigure}

\vspace{2mm}

\begin{subfigure}{\linewidth}
    \centering
    \includegraphics[width=\linewidth]{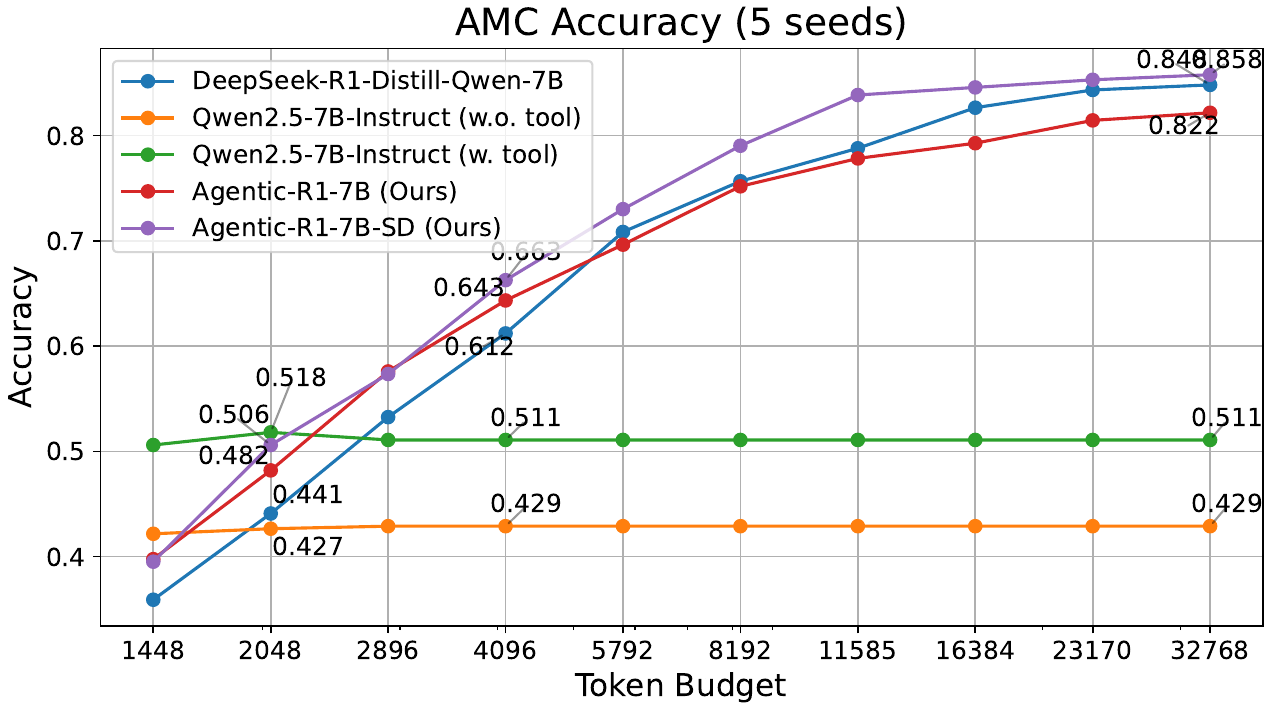}
\end{subfigure}

\caption{Performance trend of different tested models under various token budgets on the five benchmarks.}
\label{fig.full_result}
\end{figure}

We report the performance trend of different models tested in various token budgets. Please refer to Fig.~\ref{fig.full_result} for individual benchmarks and Fig.~\ref{fig.micro_avg_result} for the average.

\begin{figure*}[!ht] 
\vspace{-5mm}
\begin{AIbox}{Agentic Trajectory Example}
\small
{\color{blue}\bf User:}
{
For real numbers $x$ and $y$, define $\nabla(x,y)=x-\dfrac{1}{y}.$ Evaluate $\underbrace{\nabla(2, \nabla(2, \nabla(2, \ldots \nabla(2,\nabla(2, 2)) \ldots)))}_{2016 \,\nabla\text{s}} = \dfrac{m}{n}$ for relatively prime positive integers $m$ and $n$, and compute $100m + n$.
\\
\\
}
{\color{brown}\bf Model:} \\
{
\textit{<think>}

Let me help solve this step by step.

1) Let's understand what we're calculating:

   * We have a function $\nabla(x,y) = x - \frac{1}{y}$
   
   * We're nesting this function 2016 times
   
   * Each time, the first argument is 2
   
   * The innermost y starts with 2
   
2) Let's try to see if there's a pattern by calculating the first few terms: Let's write a Python program to help us:

\textit{</think>}

\textit{<code>}
\begin{lstlisting}[style=prompt]
def nabla(x, y):
    return x - 1/y
# Let's calculate first several terms
result = 2  # Starting value
for i in range(10):  # First 10 iterations
    result = nabla(2, result)
    print(f\"After {i+1} iterations: {result}\")
\end{lstlisting}
\textit{</code>}

\textit{<executor>}
\begin{lstlisting}[style=prompt]
After 1 iterations: 1.5
... (omitted) ...
After 10 iterations: 1.0909090909090908
\end{lstlisting}
\textit{</executor>}

\textit{<think>}

Looking at the pattern, it seems the values are converging. Let's try to understand what's happening:

1) Let's call the sequence $a_n$ where $a_1 = 2$ and $a_{n+1} = \nabla(2, a_n)$

2) For any term after the first:
   $a_{n+1} = 2 - \frac{1}{a_n}$
   
3) If this sequence converges to some value $L$, then:
   $L = 2 - \frac{1}{L}$
   
4) This gives us:
   $L^2 = 2L - 1,$
   $L^2 - 2L + 1 = 0,$
   $(L - 1)^2 = 0,$
   $L = 1$
   
5) Let's verify this is indeed what we get after 2016 iterations:

\textit{</think>}

\textit{<code>}
\begin{lstlisting}[style=prompt]
result = 2  # Starting value
for i in range(2016):
    result = nabla(2, result)
print(f\"After 2016 iterations: {result}\")
# Let's convert to fraction
from fractions import Fraction
final = Fraction(result).limit_denominator()
print(f\"\
As a fraction: {final}\")
# Calculate 100m + n
m = final.numerator
n = final.denominator
print(f\"\
100m + n = {100*m + n}\")
\end{lstlisting}
\textit{</code>}

\textit{<executor>}
\begin{lstlisting}[style=prompt]
After 2016 iterations: 1.0004957858205237
As a fraction: 2018/2017
100m + n = 203817
\end{lstlisting}
\textit{</executor>}

\textit{<think>}

After 2016 iterations, we get $\frac{2018}{2017}$, and $100m + n = 203817$. Let's verify these numbers are correct:

1) 2018 and 2017 are indeed relatively prime (2017 is prime)

2) The value is very close to but slightly larger than 1, which matches our convergence analysis

3) For $100m + n$: $100(2018) + 2017 = 203817$

Therefore, $\boxed{203817}$

\textit{</think>}

\textit{<answer>}
$\boxed{203817}$
\textit{</answer>}
}
\end{AIbox} 
\vspace{-2mm}
\caption{\textbf{An example agentic trajectory derived from OpenHands.}}
\label{fig.example_agentic_trajectory}
\vspace{-5mm}
\end{figure*}
\section{Related Work}
\label{app.related_work}

\paragraph{Tool-Augmented Reasoning.}
\label{app.related_work_tool_augmented_reasoning}
Integrating external tools into the language model chain-of-thought (CoT) has substantially improved the accuracy of numerical and factual tasks. Early program-aided methods, such as PaL~\cite{gao2023pal} and PoT~\cite{chen2023programthoughtspromptingdisentangling}, demonstrated significant gains by converting reasoning steps into executable programs, thereby delegating precise computations to code interpreters. Other lines of work, including WebGPT~\cite{nakano2022webgptbrowserassistedquestionansweringhuman} and ReAct~\cite{yao2023reactsynergizingreasoningacting}, introduced agent-like reasoning frameworks that interleave tool invocation (e.g., web searches or API calls) within multi-step reasoning. Toolformer~\cite{schick2023toolformerlanguagemodelsteach} further generalized this approach by training language models to self-supervise API calls on various tasks such as arithmetic, translation, and retrieval. START~\cite{li2025start} and CoRT~\cite{li2025cort} use hint-based prompting to activate tool use behavior, followed by rejection sampling fine-tuning (RFT) for self-improvement. However, unlike \textit{DualDistill}, these methods typically focus on short-CoT, or primarily use prompting or heuristic-based tool invocation, lacking mechanisms to automatically balance long-term reasoning against tool use based on task complexity.

\paragraph{Long Chain-of-Thought Reasoning.}
\label{app.related_work_long_chain_of_thought_reasoning}
Recent approaches have highlighted significant performance improvements by scaling the length of the inference time chain-of-thought (CoT). GPT-o1~\cite{jaech2024openai} and DeepSeek-R1~\cite{guo2025deepseek} used outcome-driven reinforcement learning to generate extensive reasoning trajectories, substantially outperforming short-CoT baselines in complex math and reasoning benchmarks. Similarly, S1~\cite{muennighoff2025s1simpletesttimescaling} and L1~\cite{aggarwal2025l1} demonstrated scaling curves showcasing a log-linear relationship between performance and inference compute. Empirical evidence supports that an increase in inference computation can often yield more cost-effective gains than an increase in model size alone~\cite{wu2025inferencescalinglawsempirical}. However, long-CoT models frequently encounter overthinking, which is the generation of overly long reasoning that leads to redundant or incorrect outcomes, especially in tool use scenarios, a phenomenon known as the reasoning-action dilemma~\cite{cuadron2025dangeroverthinkingexaminingreasoningaction}. Our work addresses these issues by teaching a student model when to switch between internal reasoning and tool-based execution adaptively.

\paragraph{Reasoning Models with Tool-Calling.}
\label{app.related_work_reasoning_models_with_tool_calling}
Recently, some works have explored the idea of combining long-form reasoning with explicit tool invocation. R1-Searcher~\cite{song2025r1searcherincentivizingsearchcapability} and Search-R1~\cite{jin2025searchr1trainingllmsreason} introduced reinforcement-learning-based retrieval policies within reasoning loops, achieving substantial performance improvements in open-domain question-answering tasks. However, unlike these methods, \textit{DualDistill} is specifically tailored for math tasks. Similarly, ReTool~\cite{feng2025retoolreinforcementlearningstrategic} trained a reasoning model with tool use for math tasks. However, unlike these approaches that rely on expensive and unstable reinforcement learning techniques, \textit{DualDistill} is a simple distillation approach, leading to a more data-efficient and practical training setup.

\paragraph{Distillation in Large Language Models.}
\label{app.related_work_distillation_in_llms}
Knowledge distillation is widely used to transfer capabilities from larger models to smaller and more efficient models~\cite{sanh2019distilbert,hsieh2023distillingstepbystepoutperforminglarger}. Recent extensions include multi-teacher distillation frameworks, which aggregate knowledge from multiple similarly structured teachers~\cite{li-etal-2024-mode}. However, existing distillation works typically assume homogeneous teacher models or single-modal reasoning paradigms. In contrast, our proposed \textit{DualDistill} explicitly utilizes heterogeneous teacher models: One specialized in agentic tool use and the other in pure textual reasoning. \textit{DualDistill} proposes an innovative approach to composing trajectories that effectively guides the student to learn from and combine both strategies.

\paragraph{Self-Taught Learning.}
A growing line of work explores methods for language models to automatically improve from their own generated outputs or feedback, a paradigm often referred to as self-taught learning or self-distillation. Early approaches in this area explored self-taught learning with pseudo-labels, where a model generates rationales or answers on unlabeled data and subsequently fine-tunes itself on these predictions~\cite{zelikman2022star, wang2022self}. \citet{shen2022self} introduces self-distillation using the output of the previous mini-batch as soft targets, improving the consistency of the output without external teachers. In the LLM era, fine-tuning LLMs on self-generated trajectories with a rule-based verifier has been applied in many domains, including coding tasks~\cite{jiang2024ledex}, theorem proving~\cite{lin2024lean}, math tasks~\cite{kumar2024training}, and others. A concurrent work~\cite{zheng2025learning} enables self-evolving reasoning across multiple modalities and fuses them during inference.
\section{License}
Our training dataset is constructed based on existing datasets, language models, and software. The following lists the relevant resources and their corresponding licenses.
\begin{itemize}
    \item OpenHands: An open-source agent framework under the MIT License;
    \item DeepSeek-R1: An open-source language model under the MIT License;
    \item Claude-3-5-Sonnet: A commercial language model under a proprietary license; accessible via the Anthropic API and supported cloud platforms;
    \item MATH500: An open-source math dataset under the MIT License;
    \item DeepMath: An open-source math dataset under the MIT License.
\end{itemize}
All third-party resources were used following their licenses and intended use, as publicly specified. Our model and dataset are released under the MIT License.
\section{Use of AI Assistants}
For this project, we use AI assistants for paper editing (e.g., grammar, spelling, word choice) and to assist with the running of experiments (e.g., scripting, automation support).
\section{Potential Risks}
Our work focuses on a model distillation framework for mathematical reasoning tasks, using synthetic and publicly available data. It does not involve deployment or the use of personal data. Therefore, we do not identify significant risks related to safety, privacy, fairness, or security within the current scope.

One potential data-related risk arises from the distillation of the teacher models. Although our training data is public, trajectory distillation may produce misleading or hallucinatory content. Such distilled data may require filtering if applied to sensitive domains in future work.
\begin{figure*}[!ht] 
\vspace{-5mm}
\begin{AIbox}{Agentic-R1 Behavior I}
\small
{\color{blue}\bf User:}
{
Evaluate the limit: \[ \lim_{n \to \infty} \sqrt[n]{n^{(4n)}+(4n)^n}\left[\left(2+\frac{1}{n^2}\right)^{18}-\left(4+\frac{4}{n^2}\right)^9\right] \]
\\
}
{\color{brown}\bf Model:} \\
{
\textit{<think>}

Let me break this down and solve it step by step.

... (Omitted) ...

Let me write some code to help verify these observations.

\textit{</think>}

\textit{<code>}
\begin{lstlisting}[style=prompt]
import numpy as np
def first_expr(n):
    return (n**(4*n) + (4*n)**n)**(1/n)
def second_expr(n):
    return (2 + 1/n**2)**18 - (4 + 4/n**2)**9
# Let's test for large n
n = 1000
print(f"First expression for n={n}: {first_expr(n)}")
print(f"Second expression for n={n}: {second_expr(n)}")
\end{lstlisting}
\textit{</code>}

\textit{<executor>}
\begin{lstlisting}[style=prompt]
OverflowError: int too large to convert to float
\end{lstlisting}
\textit{</executor>}

<think>

Let me modify the code to use logarithms to better understand the growth:

</think>

\textit{<code>}
\begin{lstlisting}[style=prompt]
... (Omitted) ...
print(f"Log of first expression grows like {log_first_expr(1000)/1000}")
print(f"Log of second expression grows like {log_second_expr(1000)}")
\end{lstlisting}
\textit{</code>}

\textit{<executor>}
\begin{lstlisting}[style=prompt]
Log of first expression grows like 10.907755278982137
Log of second expression grows like 0.0
\end{lstlisting}
\textit{</executor>}

\textit{<think>}

Let me analyze this:

... (Omitted) ...

Let me verify this with very large n:

\textit{</think>}

\textit{<code>}
\begin{lstlisting}[style=prompt]
n = 10**10
first = (n**(4*n) + (4*n)**n)**(1/n)
second = (2 + 1/n**2)**18 - (4 + 4/n**2)**9
print(f"First expression = {first}") # This will show as inf, but we know it grows like n^4
print(f"Second expression = {second}")  # Should approach 0
\end{lstlisting}
\textit{</code>}

\textit{<executor>}
\begin{lstlisting}[style=prompt]
Error: Code execution timed out.
\end{lstlisting}
\textit{</executor>}

\textit{<think>}

... (Omitted) ...

{\color{red}\textbf{Wait, the code is not correct, let's try text reasoning.}}

Okay, so I need to evaluate this limit: 

... (Omitted) ...

However, 2\textasciicircum 16 is 65536, and 65536 * 9 is indeed 589,824. So, perhaps the answer is 589,824.

\textit{</think>}

\textit{<answer>}
$\boxed{589,824}$
\textit{</answer>}
}
\end{AIbox} 
\vspace{-2mm}
\caption{\textbf{Example Behavior I of \textit{Agentic-R1}.} The model initially attempts to solve the problem using the tool to compute a limit expression, but fails after several attempts. Notably, it then transitions to a pure text-based reasoning strategy and successfully arrives at the correct solution.}
\label{fig.agentic-r1_behavior_1}
\vspace{-5mm}
\end{figure*}
\begin{figure*}[!ht] 
\vspace{-5mm}
\begin{AIbox}{Agentic-R1 Behavior II}
\small
{\color{blue}\bf User:}
{
Determine the number of tuples \((k_1, k_2, \ldots, k_{26})\) such that each \(k_i\) is either 0, 1, or 3, and the sum \(k_1 + k_2 + \ldots + k_{26} = 15\).
\\
}
{\color{brown}\bf Model:} \\
{
\textit{<think>}

Okay, so I need to figure out how many 26-tuples there are where each element is either 0, 1, or 3, and the total sum is 15.

... (Omitted) ...

Let me compute each term step by step.

First, y=0:

x=15, y=0. Then the number of tuples is C(26,15)*C(11,0). C(26,15) is the same as C(26,11) because C(n,k) = C(n, n-k). So C(26,11). C(11,0) is 1. So total is C(26,11).

Similarly, y=1:

x=12, y=1. Number of tuples is C(26,12)*C(14,1). C(26,12) is the same as C(26,14). C(14,1) is 14. So total is 14*C(26,14).

{\color{red}\textbf{Wait, use text reasoning is too tedious, let's try code reasoning.}}

\textit{</think>}

\textit{<think>}

Let me think about this problem systematically.

Let's break this down:

1) We need to find the number of 26-tuples where each element is 0, 1, or 3

2) The sum of all elements must be 15

3) This is a constrained counting problem that can be solved using dynamic programming

4) Let's use Python to implement a solution

5) We'll need to:

\ \ \ \  - Create a 2D DP table where dp[i][j] represents the number of ways to make sum j using first i numbers
   
\ \ \ \  - For each position i, we can use 0, 1, or 3
   
\ \ \ \  - We'll build this up from 0 to 26 numbers
   
\ \ \ \  - The final answer will be in dp[26][15]

\textit{</think>}

\textit{<code>}
\begin{lstlisting}[style=prompt]
def count_tuples():
    # dp[i][j] represents number of ways to make sum j using first i numbers
    dp = [[0] * 16 for _ in range(27)]  # 27 numbers (0-26) and sums 0-15
    # Base case: 0 numbers can only make sum 0
    dp[0][0] = 1
    for i in range(1, 27):
        for j in range(16):
            if j >= 0: dp[i][j] += dp[i-1][j] # Try adding 0
            if j >= 1: dp[i][j] += dp[i-1][j-1] # Try adding 1
            if j >= 3: dp[i][j] += dp[i-1][j-3] # Try adding 3
    return dp[26][15]
result = count_tuples()
print(f"The number of valid 26-tuples is: {result}")
\end{lstlisting}
\textit{</code>}

\textit{<executor>}
\begin{lstlisting}[style=prompt]
The number of valid 26-tuples is: 853423740
\end{lstlisting}
\textit{</executor>}

\textit{<think>}

Let me verify this result with another approach to make sure it's correct:

\textit{</think>}

\textit{<code>}
\begin{lstlisting}[style=prompt]
... (Omitted) ...
verified_result = verify_result()
print(f"Verified result: {verified_result}")
\end{lstlisting}
\textit{</code>}

\textit{<executor>}
\begin{lstlisting}[style=prompt]
Verified result: 853423740
\end{lstlisting}
\textit{</executor>}

\textit{<think>}

I have verified the result using two different approaches and both give the same answer.

\textit{</think>}

\textit{<answer>}
$\boxed{853,423,740}$
\textit{</answer>}
}
\end{AIbox} 
\vspace{-2mm}
\caption{\textbf{Example Behavior II of \textit{Agentic-R1}.} The model first attempts to solve the combinatorial problem via text-based reasoning, but switches to tool-assisted reasoning due to computational complexity. It then implements a dynamic programming algorithm to solve the problem efficiently and verify the result.}
\label{fig.agentic-r1_behavior_2}
\vspace{-5mm}
\end{figure*}

\end{document}